\def\year{2020}\relax
\documentclass[letterpaper]{article}
\usepackage{aaai21}
\usepackage{times} 
\usepackage{helvet} 
\usepackage{courier} 
\usepackage[hyphens]{url} 
\usepackage{graphicx} 
\usepackage{natbib}  
\usepackage{caption}
\usepackage{microtype}
\usepackage{subfigure}
\usepackage{booktabs} 
\usepackage{amssymb}
\usepackage{amsmath}
{
      \newtheorem{assumption}{Assumption}
  }
\usepackage{algorithm}
\usepackage[noend]{algpseudocode}
\usepackage{enumitem}
\usepackage{multirow}
\usepackage{siunitx}
\newcommand{\comment}[1]{}
\newcommand{\argmin}{\mathop{\mathrm{argmin}}}
\newcommand{\argmax}{\mathop{\mathrm{argmax}}}
\newcommand{\ts}{\textsuperscript}
\def\BState{\State\hskip-\ALG@thistlm}

\usepackage{hyperref}

\title{Overcoming Model Bias for Robust Offline Deep Reinforcement Learning}

\author{\Large \textbf{Phillip Swazinna\textsuperscript{\rm 1,2}, Steffen Udluft\textsuperscript{\rm 1}, Thomas Runkler\textsuperscript{\rm 1,2}}\\}

\affiliations{
\textsuperscript{\rm 1}Siemens Corporate Technology, Munich, Bavaria, Germany\\
\textsuperscript{\rm 2}Department of Informatics, Technical University of Munich, Munich, Bavaria, Germany \\
swazinna@in.tum.de
}

\begin{document}
\maketitle

\begin{abstract}
State-of-the-art reinforcement learning algorithms mostly rely on being allowed to directly interact with their environment to collect millions of observations. This makes it hard to transfer their success to industrial control problems, where simulations are often very costly or do not exist, and exploring in the real environment can potentially lead to catastrophic events. Recently developed, model-free, offline RL algorithms, can learn from a single dataset (containing limited exploration) by mitigating extrapolation error in value functions. However, the robustness of the training process is still comparatively low, a problem known from methods using value functions. To improve robustness and stability of the learning process, we use dynamics models to assess policy performance instead of value functions, resulting in MOOSE (MOdel-based Offline policy Search with Ensembles), an algorithm which ensures low model bias by keeping the policy within the support of the data. We compare MOOSE with state-of-the-art model-free, offline RL algorithms { BRAC,} BEAR and BCQ on the Industrial Benchmark and MuJoCo continuous control tasks in terms of robust performance, and find that MOOSE outperforms its model-free counterparts in almost all considered cases, often even by far.
\end{abstract}

\section{Introduction}
\label{intro}
In reinforcement learning (RL), the goal is to train an agent, which will through interactions with its environment maximize a utility value referred to as reward. Algorithms that train such an agent must usually carefully balance between exploring their environment in order to increase their knowledge about it, or exploiting their knowledge to achieve the highest rewards possible \citep{sutton1998introduction}. The ability to explore is fundamental to the idea of reinforcement learning, and questions such as when and how to explore efficiently and effectively play a big role in reinforcement learning research today \citep{schmidhuber2006developmental,bellemare2016unifying,osband2019deep}. However, in this paper we consider the case where no exploration is possible at all, as that is a widespread constraint in practice, which is often overlooked in literature.

\begin{figure}[h]
    \center{\includegraphics[width=0.5\textwidth]
    {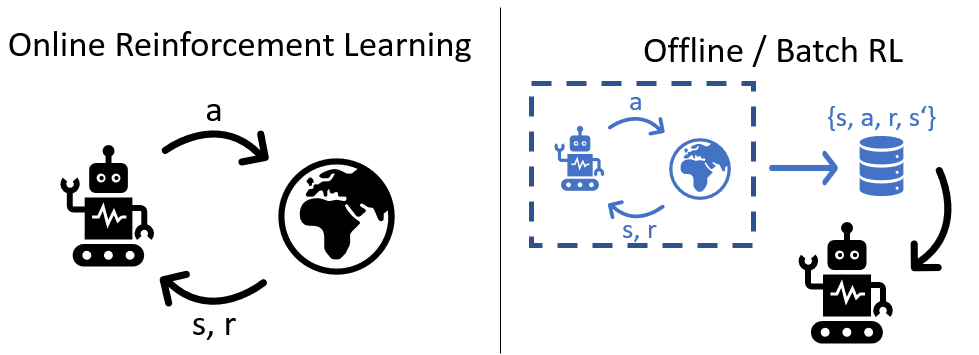}}
    \caption{\label{fig:rlsettings} Difference between commonly assumed online RL, where the learning agent directly interacts with its environment, versus offline RL, where interaction is impossible since we only have access to data instead of the environment.}
\end{figure}

\begin{figure}[h]
    \center{\includegraphics[width=0.45\textwidth]
    {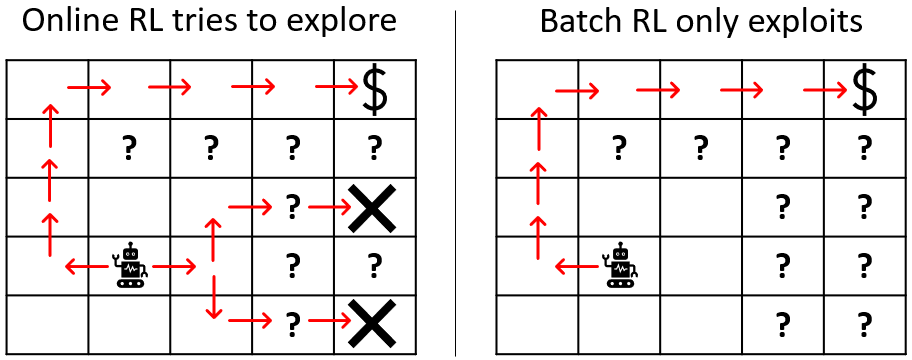}}
    \caption{\label{fig:exploreexploit} Online RL methods generally cannot be used in offline RL settings, since they have been found to break when on-policy or close to on-policy observations are unavailable \citep{fujimoto2018off}. Offline RL methods on the other hand do not require on-policy data and simply exploit the knowledge they gained from the initial dataset.}
\end{figure}

By combining classic reinforcement learning techniques with modern function approximation, recent RL algorithms have managed to achieve tremendous results in a variety of domains, such as video games, robot locomotion, and other continuous control tasks \citep{mnih2013playing,lillicrap2015continuous,schulman2015trust,schulman2017proximal,haarnoja2018soft}. Most of these approaches belong to the family of online, model-free, actor-critic methods. They alternate between updating the neural agent via policy gradient, and using it to collect more observations \citep{williams1992simple,schnegass2007improving,silver2014deterministic,haarnoja2018soft}.
Recently, it has been found that they can only learn from data that has been collected under the current or close to the current policy, even if they come with the "off-policy" attribute \citep{fujimoto2018off}. They can thus only learn from live interactions with an environment as opposed to learning from a previously collected dataset, as they constantly need to collect on-policy data. This makes it hard to transfer the success that RL methods have had to settings frequently encountered in real-world applications, where large datasets have been collected passively via logging (i.e., turbine or factory control, autonomous vehicles, etc.), and opportunities to collect on-policy data (explore) are missing, since doing so could be dangerous, or simply prohibitively expensive.\\
In this paper we thus consider learning an agent without being allowed any direct interaction with the environment, resulting in no possibility for exploration. The agent training has to be based solely on a single previously collected batch of interactions which is provided up front. Since it is unclear how that batch has been generated, the training needs to work with datasets containing different levels of (or even no) exploration. By considering this so-called offline \citep{lange2012batch}, or batch RL setting, we seek to move RL closer to real world applications since it is a classic constraint in industrial machine learning rarely addressed by RL literature. Figures \ref{fig:rlsettings} and \ref{fig:exploreexploit} visualize the differences between online RL, where exploration is possible, and offline RL, where it is not.\\
Recently, algorithms have been developed that explicitly address the issue of extrapolation error due to missing support in the dataset \citep{fujimoto2018off,kumar2019stabilizing}, however most other offline algorithms often implicitly assume that the data in the batch contains sufficient exploration to solve the problem \citep{ernst2005approximate,depeweg2016learning,hein2016reinforcement,hein2018interpretable}. In reality this assumption is likely violated, since datasets are usually collected without being explicitly designated to be used in a reinforcement learning setting. It may even be unclear how the data was collected (human interactions,   {classic controllers, policies derived by current RL algorithms}, mixtures). Hence, we look for algorithms which perform as well as possible, given the amount and quality of exploration contained in the batch.\\
The robustness with which policies are produced is arguably the most important factor in offline RL, since without environment access, we have no reliable way to perform policy selection. As model-based RL is often attributed superior sample efficiency and greater stability compared to model-free methods, we find it much more suitable in the context of the innately limited-data scenario that is offline reinforcement learning. Hence, we develop a model-based offline RL algorithm that is otherwise closely related to state-of-the-art model-free offline algorithms { BRAC,} BEAR and BCQ { \citep{fujimoto2018off,kumar2019stabilizing,wu2019behavior}}, in order to investigate whether model-based RL can play to its strengths in this setting.

\section{Related Work}
\label{related}
Early work in batch reinforcement learning includes least squares policy iteration (LSPI) \citep{lagoudakis2003least}, which directly embeds itself in the policy iteration framework \citep{sutton1998introduction}, fitted Q iteration (FQI) \citep{ernst2005treebased}, which augments the batch with samples where the reward is computed based on the temporal difference Q target, as well as its neural network based counterpart, neural fitted Q iteration (NFQ) \citep{riedmiller2005neural}. While all three algorithms are theoretically able to work in the \textit{pure} batch mode, also called offline reinforcement learning, NFQ is usually referred to as being a semi-batch, or growing batch algorithm, since it is usually used to collect more data as the Q-function estimate changes over time \citep{lange2012batch}, enabling it to solve rather complex problems such as playing robot soccer \citep{riedmiller2009reinforcement}. It is thus more closely related to current state-of-the-art online RL algorithms, which employ experience replay as a measure to increase sample efficiency \citep{adam2011experience,wang2016sample,andrychowicz2017hindsight}. Early pure batch algorithms have been mainly used to solve rather simple MDPs and / or contain the implicit assumption that the batch contains a balanced set of transition samples spanning the entire state-action space \citep{kalyanakrishnan2007batch}.\\ \\
Behavior cloning methods \citep{ng2000algorithms,ross2010efficient,ho2016generative,laskey2017dart,nair2018overcoming,codevilla2019exploring} constitute a step towards learning directly from data: They learn to imitate an expert policy, usually from trajectories that were obtained by applying the expert policy in an environment. Plain behavior cloning without further environment interaction can thus be seen as a special case of offline reinforcement learning, where we have additional information about the behavior policy. These methods face difficulties once the trained policy leaves the known region of the state-action space on the real system---a problem referred to as covariate shift \citep{ho2016generative,codevilla2019exploring}. One way to deal with it, is by injecting noise into the expert policy, thereby forcing the expert to demonstrate how to recover when things go wrong \citep{laskey2017dart}. Another path is inverse reinforcement learning, where algorithms first learn to infer a reward function from expert demonstrations, and then have an inner reinforcement learning loop to solve it \citep{ng2000algorithms}. Behavior cloning can also be simply used as a way to warm start policies, after which online reinforcement learning can start \citep{nair2018overcoming}. All these methods provide an opportunity to significantly reduce the amount of environment interactions needed, with the crucial caveat that they all assume the availability of (data generated by) an expert, which is in practice often unavailable.\\ \\
Batch constrained Q-learning (BCQ) \citep{fujimoto2018off} examines the setting where the agent is only provided with a single batch of off-policy data, that may contain sub-optimal exploration or even no exploration at all, and in contrast to imitation learning techniques also may or may not contain expert demonstrations. While traditional Q-learning approaches would fail in this context due to wrong extrapolation values in areas of the state-action space which have not been explored, BCQ augments methods from behavior cloning to constrain its policy to only select state-action pairs which are close to the ones encountered in the original batch. Bootstrapping error accumulation reduction (BEAR) \citep{kumar2019stabilizing}, { as well as behavior regularized actor critic (BRAC) \citep{wu2019behavior}} pick up on this idea and embed it in the actor-critic paradigm by learning a closed form policy which is constrained by the maximum mean discrepancy \citep{gretton2012kernel} or the KL divergence between likely actions and the recommended action of the policy, respectively. \citet{siegel2020keep} find adaptive behavioral priors to develop a model-free offline RL framework which is constrained to stay close only to actions that have a value higher or equal to those currently proposed by the policy. \citet{leebatch} address the issue of finding the right hyperparameters for offline RL algorithms. While these methods have been developed for continuous state and action spaces, even more algorithms have been developed for the offline setting with discrete actions \citep{laroche2017safe,dabney2018implicit,agarwal2019striving,jaques2019way,fujimoto2019benchmarking}.\\ \\
Model-based reinforcement learning has shown early on that it can increase data efficiency over value-function based RL by explicitly learning a transition model \citep{sutton1990integrated}. This advantage over model-free reinforcement learning has been demonstrated many times over \citep{deisenroth2011pilco,kurutach2018model,nagabandi2018neural}; however, it usually comes with the downside of lower asymptotic performance due to model bias \citep{nagabandi2018neural,pong2018temporal}. Policies trained on imperfect models can diverge by accumulating transition errors or extrapolate falsely and lead policies to favor visiting parts of the state-action space in which models are incorrect and overly optimistic. { It has been shown however, that this issue can, even in the offline RL setting, usually be circumvented by penalizing model uncertainty in the policy training process, e.g., by employing Bayesian models \citep{depeweg2016learning,depeweg2017decomposition,kaiser2020bayesian}, or by ensembling, as in MOPO and MOReL \citep{yu2020mopo,kidambi2020morel}. MOPO facilitates two head architecture models that predict mean and variance of successor states and subtracts the maximum variance across a model ensemble from the predicted mean reward. MOReL on the other hand uses unknown state detectors based on model ensemble disagreement to end trajectories when successor states become too uncertain. Both algorithms then combine the model ensemble in a dyna fashion with Q-function based agents, such as PPO or TRPO\citep{schulman2015trust, schulman2017proximal}, to plan through the newly generated data. These approaches to offline RL are thus related to ours only in the sense that they are model-based, however in this work we neither follow any model-uncertainty based approaches to keep the policy from visiting unfavorable regions of the state-action space, nor do we use any value-functions. Instead, we show that a behavior regularization based approach more similar to the one employed by { BRAC, BEAR, and BCQ}, can be transferred to a purely model-based setting in order to reap the benefits of both approaches: Being constrained to stay in the support of the data as well as better sample efficiency and stability.}

\section{MOdel-based Offline policy Search with Ensembles (MOOSE)}
\label{MOOSE}
We are interested in applying reinforcement learning in real-world problems which exhibit complex environment dynamics, high dimensional continuous state and action spaces and complicated noise patterns, where we are only given a single batch of data and are not allowed to collect any further observations---as a good example we may consider turbine control. We explicitly make no assumption on the way in which the data was generated. It may thus be, that we are dealing with data generated by expert policies, suboptimal policies, or even human controllers. Furthermore, we consider practical aspects of the algorithms: Since final policies need to be deployed in live systems, stochastic policies (e.g., SAC, BCQ \cite{haarnoja2018soft,fujimoto2018off}) require additional provisions for safe operation and we thus do not consider them. Due to the data scarcity innate to the problem definition, sample efficiency is crucial and we thus design a model-based algorithm. This furthermore facilitates monitoring, as well as the input of prior domain knowledge by human experts, which may be of critical interest in real-world reinforcement learning problems.\\ \\
In this section, we develop a model-based offline RL algorithm that works in continuous state-action spaces and can handle arbitrarily generated batches of data while still improving upon the behavior policy, by constraining the trained policy to stay close to state-action pairs in the batch. Our approach is neither based on sampling actions from a generative model (like e.g., BEAR \& BCQ \citep{fujimoto2018off,kumar2019stabilizing}), nor does it constrain the trained policy directly to be close to the generating policy (like KL-control, SPIBB-DQN \citep{jaques2019way,laroche2017safe}). Instead, it penalizes state-action pairs unlikely under the generative policy. We show that our method beats state-of-the-art model-free offline algorithms, especially in terms of robustness, which is arguably the most important aspect when considering policy deployments in the real world.

\subsection{Model Training and Standard Model-Based Reinforcement Learning}
\label{sec:mbrl}
We assume to be given a batch of transition samples $\{(s_t, a_t, r_t, s_{t+1}) | t = 0 \dots T-1\}$, where $T$ is the number of steps the behavior policy(ies) was running (possibly over multiple trajectories). In order to make most effective use of the information, we train a neural network model $f$, parameterized by $\phi$, to represent the environment's transition dynamics $s_{t+1} = f_{\phi}(s_t, a_t)$ and possibly its reward function. We can also accommodate à priori known reward functions (i.e., $r_t = r(s_t, a_t, s_{t+1})$, where we assume $r$ to be differentiable). We apply commonly used techniques for model learning, such as weight normalization \citep{salimans2016weight} and normalizing training data to have zero mean and unit standard deviation. In some environments it can be beneficial to model the transition difference $\Delta s = s_{t+1} - s_t$ instead of directly predicting $s_{t+t}$. The training loss can thus either be given by the mean squared error between predicted delta and true delta (1) or between predicted state and true state (2) and is optimized via Adam \citep{kingma2014adam}. 
\begin{eqnarray}
\label{eq:model_train}
    L(\phi) &=& \sum_t ||f_{\phi}(s_t, a_t) - \frac{(s_{t+1} - s_t) - \mathbf{\mu^{\Delta s}}}{\mathbf{\sigma^{\Delta s}}}||_2 \\ 
    &&or \nonumber \\
    L(\phi) &=& \sum_t ||f_{\phi}(\frac{s_t - \mathbf{\mu^{s}}}{\mathbf{\sigma^{s}}}, a_t) - \frac{s_{t+1} - \mathbf{\mu^{s}}}{\mathbf{\sigma^{s}}}||_2
\end{eqnarray}

With slight abuse of notation, we will denote the estimated next state as $\hat{s}_{t+1} = f_{\phi}(s_t, a_t)$. With the differentiable neural network transition (and reward) model, we then derive a standard actor-critic training algorithm: We train a neural network based policy $\pi$ with parameters $\theta$ by assessing its performance using imagined trajectories generated by rolling out the model into the future. Since stochastic behavior can be inappropriate in real-world, safety critical systems, we assume the policy to be deterministic, i.e., $a = \pi_\theta(s)$. The expected cumulative discounted return of the policy is then estimated using $N$ rollouts of horizon $H$. We improve the quality of the return estimate by using an ensemble of $K$ transition models. Its negated value can then be minimized in order to optimize the policy's parameters $\theta$:

\begin{align}
\label{naive_mbrl}
    L(\theta) =& - \frac{1}{KN}\sum_k \sum_n \sum_t \gamma^t r(s_t, \pi_\theta(s_t), f_{\phi_k}(s_t, \pi_\theta(s_t))) \nonumber \\
    =& - \mathbb{E}_{\pi, f_{1..K}}[R]
\end{align}
We sample $s_0$ from the start states in the dataset, $s_t = f_{\phi_k}(s_{t-1}, \pi_\theta(s_{t-1}))$ and $r(s_t, a_t, s_{t+1})$ is either a learned or an à priori known reward function.\\
If we were to assume infinite amounts of perfectly explored data, or at least continuous collection of further observations under the trained policy, no further adjustments to our approach would be necessary. However, since we do not know which parts of the state-action space have been explored to a level of confidence, we risk unjustified predictions by the transition model in unexplored regions. The policy may intentionally try to exploit the erroneous reward estimates predicted in those regions, resulting in great imagined policy performance, but very poor performance once deployed in the real system. In the following, we thus introduce an approach to constrain the trained policy to refrain from visiting these regions, mitigating a shortcoming often attributed to model-based RL algorithms: Increased bias compared to model-free methods.

\subsection{Reducing Model Bias in Model-Based Reinforcement Learning}
\label{sec:reduce_bias}

We would like to reduce the visitation of state-action pairs by the trained policy, for which we cannot accurately assess the transition to the next state (and consequently also the reward), since it would lead to inaccurate estimates of the policy's expected performance in Equation \ref{naive_mbrl}. By constraining the trained policy to stay close to the known region of state-action pairs, we aim to minimize this error or bias of the transition models, since it can be assumed their predictions will be more accurate in regions close to the batch data distribution.\\
We quantify model bias in transition steps of trajectories, as being the expected state prediction error, when comparing real transitions $(s, a, r, s')$ produced by the true environment $e$ with virtual transitions $(\hat{s}, \hat{a} = \pi(\hat{s}), \hat{r}, \hat{s}')$ from rollouts through the learned model $f$, where trajectories started in the same state $s_0$ and where the timestep $t$ is equal for both $s$ and $\hat{s}$.
\begin{eqnarray}
      b &   = &   \mathbb{E}_{s' \sim e(s,a),\; \hat{s}'=f(\hat{s},\hat{a})} \left[||s' - \hat{s}'||_2\right]\\
    &  =&   \mathbb{E} \left[||e(s,a) - f(\hat{s}, \hat{a})||_2 \right] \nonumber
\end{eqnarray}{}
{ 
Note that while the trained dynamics models are deterministic, i.e., $\hat{s}'=f(\hat{s}, \hat{a})$, this is not necessarily true in the real environment, so that successor states may follow complex distributions, which is why we denote $s' \sim e(s,a)$.}\\
In order to train a well performing policy, in addition to maximizing expected virtual rewards, we would like to minimize the expected value of the model's bias throughout trajectories, since otherwise we will not be able to accurately assess the policy's quality. With $\mu_{\pi}^e(s)$ being the state visitation probability of some policy $\pi$ in the actual environment (and $\mu_{\pi}^f(s)$ being the corresponding estimated probability under the model), we write the expected value of the bias as
\begin{equation}
    \mathbb{E}_{s \sim \mu_{\pi}^e, a \sim \pi(s), \hat{s} \sim \mu_{\pi}^f, \hat{a} \sim \pi(\hat{s})} \left[B\right]
\end{equation}{}
where $B$ is the sum of biases accumulated throughout a trajectory in a model:
\begin{equation}
    B = \sum_t b_t = \sum_t {  \mathbb{E}_{s_{t+1} \sim e(s_t,a_t)} \left[ ||s_{t+1} - f(\hat{s}_t, \pi(\hat{s}_t))||_2 \right]}
\end{equation}

We cannot compute the expectation of $B$ for arbitrary policies $\pi$ though, as we have no access to the true environment $e$ and have no way of estimating $\mu_\pi^e$. Since we are only ever given a single batch of data and are not allowed to try new policies in the actual environment, the only state visitation probability distribution that we can estimate under the true environment is the one for the behavior policy(ies) $\beta$ that generated the batch. The expected model bias for a newly trained policy $\pi$ is then:
\begin{equation}
    \mathbb{E}_{s \sim \mu_{\beta}^e, a \sim \beta(s), \hat{s} \sim \mu_{\pi}^f, \hat{a} \sim \pi(\hat{s})} \left[B\right]
\end{equation}{}

When training a policy $\pi$, we would like $B$ to be low (i.e., below some threshold close to zero) in order to be able to accurately assess its performance. It is intuitively clear that constraining the trained policy to be close to the behavior policy, i.e., in terms of the KL divergence (like \cite{laroche2017safe, wu2019behavior}) could be an adequate solution since minimizing
\begin{equation}
    \theta^* = \argmin_\theta \rm{KL}(\beta(a|s)||\pi_\theta(a|s))
\end{equation}
would move $\pi$ close to $\beta$, and consequently $\mu_\pi^e(s)$ would be close to $\mu_\beta^e(s)$ as well as to $\mu_\pi^f$, and thus, $B$ is likely to be low since the model(s) can well predict outcomes of the policy's actions. We would however like to point out, that this solution is \textbf{(a)} not optimal in terms of bias reduction, since even low probability behavior of the original policy can be copied, which is likely not as well predictable by a transition model as actions that are selected with higher probability, and \textbf{(b)} possibly too restrictive since it will be hard to outperform a policy that you are trying to mimic closely. Furthermore, the above approach is simply not feasible since we aim for deterministic policies.\\
However, it is clear that in order to have low bias given some action $\hat{a}$ by the policy $\pi$ while being in the imagined state $\hat{s}$, both:
\begin{itemize}
    \item the true visitation probability of this state $\hat{s}$ under the behavior policy, i.e., $\mu_{\beta}^e(\hat{s})$
    \item as well as the true action selection probability under the behavior policy, i.e., $\beta(\hat{a}|\hat{s})$
\end{itemize}{}
should be as large as possible, or at least above a certain threshold. Otherwise, we either estimated to be in a state which was rarely visited under the behavior policy (which likely happened because we took an action unlikely under the behavior policy) or are taking an action in this state which was rarely executed, leading to a large probability that the model predicts an incorrect successor state.\\
As the only thing we are allowed to change in this setting are the parameters $\theta$ of the policy $\pi$ that we are training, we would like to constrain them in order to allow a large probability $\mu_{\pi}^f(\hat{s})\pi(\hat{a}|\hat{s})$ only when $\mu_{\beta}^e(\hat{s})\beta(\hat{a}|\hat{s})$ is also high.\\
{ To write down our intuition mathematically, we need to make an assumption about how the behavioral policy's probability of generating a state-action pair $(s, a)$ in the real environment (i.e., having seen $(s, a)$ in our dataset) influences the magnitude of the error our trained transition models typically make in predicting the corresponding successor state.}
\begin{assumption}
    { 
    The distribution of model errors $e(s, a) - f(s, a)$ has a variance that is monotonically decreasing with the probability of having seen the imagined data in reality, i.e., that the data sample $(s, a)$ was generated under the original environment dynamics $e$ and the behavior policy $\beta$:}
    \begin{equation}
        (s' - \hat{s'}) \sim \mathcal{N}(0, -\log p_{e,\beta}(s, a))
    \end{equation}
\end{assumption}
{ Minimizing expected bias then nicely corresponds to maximizing state and action probability under the behavior policy and real environment, since the expectation of a squared Gaussian variable is its variance. See Appendix \ref{app_eq_minimize} for a more detailed derivation.}
\begin{eqnarray}
\label{eq_minimize}
    \mathbf{\theta^*} =& \argmin\limits_\theta & \mathbb{E}[B]\\
     =&  \argmin\limits_\theta &  \mathbb{E}[(s' - \hat{s'})^2] \nonumber\\
     =&  \argmin\limits_\theta &  \mathbb{E}[-\log p_{e,\beta}(\hat{s}, \hat{a})] \nonumber\\
     =&  \argmax\limits_\theta &  \mathbb{E}[p_{e,\beta}(\hat{s}, \hat{a})] \nonumber\\
    =& \argmax\limits_\theta & \mathbb{E} [\mu_{\beta}^e(\hat{s})\beta(\hat{a}|\hat{s})] \nonumber
\end{eqnarray}{}
{ 
We would like to point out that another difference between our method and a simple KL regularization of the policy to be close to the behavioral one is, that we not only reward closeness in terms of selecting similar actions when being in the same state, but moreover we also reward state visitation when the state is close to a state seen in the batch.} While the former implies the latter when policies are regularized so strongly that they eventually become identical, we find this detail important, because as previously mentioned, we would like the policy to not be regularized too strictly, as long as it does not move outside the region of known states. \\

\subsection{A Practical Algorithm}
Following the ideas to reduce model bias for a model-based, offline reinforcement learning algorithm in the previous sections, we need a way to estimate the state-visitation and action selection probability of a state-action pair under the behavioral policy in the original environment. Since we neither assume the actual environment $e$, nor the behavior policy $\beta$ to be given, we will in the following approximate $\mu_{\beta}^e(\hat{s})\beta(\hat{a}|\hat{s})$ using a variational autoencoder (VAE) \citep{kingma2013auto} $v$, parameterized with weights $\omega$. {  The variational autoencoder aims to model the probability of the data points $(s,a)$ to occur, by minimizing the evidence lower bound (ELBO), where we place a Gaussian prior on the latent variables}:

\begin{gather}
    L(\omega) = \mathbb{E}_{q_\omega(z|s,a)} [-\log p_\omega(s,a|z)] + D_{KL}(q_\omega(z|s,a)||p(z)) \nonumber \\
    p(z) \sim \mathcal{N}(0, 1) \label{eq:penalizer_train}
\end{gather}
To use it in order to constrain the policy towards having lower model bias, we { leverage that the ELBO is under standard assumptions (normally distributed modeling errors in Euclidean space, which is reasonable for the considered benchmarks) optimized using reconstruction error as given by the mean squared error (MSE). We can thus use reconstruction errors of state-action pairs as a proxy for their probability of occuring.} Low reconstruction errors will indicate that the pair was likely to be visited under the original environment and behavior policy, while large errors will indicate unlikely state-action pairs. We accumulate this penalty over the course of the imagined trajectories through the trained transition model:
\begin{equation}
\label{eq:penalty}
    \mathbb{E}[P] = \sum_t \mathbb{E}_{q_\omega(z|s,a), (s,a) \sim \pi,f} [-\log p_{\omega}(s,a|z)]
\end{equation}
and use it in a convex combination with the return estimate to penalize the policy, expanding Equation \ref{naive_mbrl}:
\begin{equation}
\label{eq:final_loss}
    L(\theta) = -\lambda \mathbb{E}[R] + (1-\lambda)\mathbb{E}[P]
\end{equation}{}
We furthermore take inspiration from double Q-learning \citep{fujimoto2018off} and generalize it for our model-based approach: In order to further avoid uncertainty over the reward estimates, we use the trained ensemble of $K$ reward models to be more conservative and bias the estimate of the return towards the minimum of the models:
\begin{eqnarray}
    \label{eq:weighted_return}
    \mathbb{E}[R] = \eta \min_k\left\{\sum_t \gamma^t r(s_t, \pi_\theta(s_t), f_k(s_t, \pi_\theta(s_t)))\right\} \\
    + (1-\eta) \frac{1}{K}\sum_k \left[ \sum_t \gamma^t r(s_t, \pi_\theta(s_t), f_k(s_t, \pi_\theta(s_t)))\right] \nonumber
\end{eqnarray}
This concludes MOOSE, where we combine model-based policy assessment with a measure of how likely state-action pairs would have been visited under the data generating policy. In an effort to mitigate model bias we train a policy to only visit state-action pairs supported by the batch, resulting in the ability to learn in offline RL settings where no additional data may be collected. Algorithm \ref{algo} summarizes MOOSE in pseudocode.

\begin{algorithm}
\caption{MOOSE}\label{algo}
\begin{algorithmic}[1]
\Procedure{MOOSEpolicy}{$D=\{s_i, a_i, s_{i+1}, r_i\}$}
    \State train dynamics models $f_{1..K}$ with $D$ and Equation \ref{eq:model_train}
    \State train VAE $v$ with $D$ and Equation \ref{eq:penalizer_train}
    \State init policy network $\pi_\theta$
    \For{\texttt{j in 1..policyupdates}}
        \State sample start states $S_0$ from $D$
        \State estimate $\mathbb{E}[R]$ using $f_{1..K}$ and Equation \ref{eq:weighted_return}
        \State estimate $\mathbb{E}[P]$ using $v$ and Equation \ref{eq:penalty}
        \State $\theta_j \gets \theta_{j-1} - \alpha \nabla_{\theta_{j-1}} \left[-\lambda \mathbb{E}[R] + (1-\lambda)\mathbb{E}[P]\right]$
    \EndFor
    \State \textbf{return $\pi_\theta$};
\EndProcedure
\end{algorithmic}
\end{algorithm}

\section{Experiments}

We perform experiments with MOOSE and state-of-the-art offline RL algorithms BEAR \& BCQ in MuJoCo environments as well as in the Industrial Benchmark to investigate:
\begin{enumerate}
    \item how the algorithms handle various continuous control environments with high dimensional state and action spaces, complicated state transitions, delayed rewards, and complex noise patterns. We aim to design a general algorithm that can perform well in various environments featuring continuous state and action spaces.
    \item how the algorithms handle varying generating policies and different degrees of exploration contained in the batch. The algorithm should not depend on the type of policy or the circumstances which generated the data. We thus perform experiments with   {structurally different} policies, deterministic and probabilistic policies, and policies resulting in narrow as well as much wider data distributions.
    \item how MOOSE compares to state-of-the-art model-free off-line algorithms. We deliberately designed our algorithm to be close to { BRAC,} BEAR and BCQ with the main difference being that it is model-based instead of model-free, since we hypothesize that due to the superior sample efficiency usually attributed to model-based RL, it is much better suited for the data-scarce nature of offline RL.
    \item the stability of the learning process. This is arguably the most important aspect of the algorithms. In supervised learning, we can do model selection via a validation set (i.e., early stopping). Since there is no equivalent technique for policy selection in offline reinforcement learning, it is crucially important that learning is robust, so that policy selection is insensitive to initializations or policy updates.
\end{enumerate}
We furthermore include a comparison to commonly known DDPG \citep{lillicrap2015continuous}, to include an algorithm which is off-policy, but not designed for the offline setting.

\subsection{MuJoCo}
MuJoCo continuous control tasks \citep{todorov2012mujoco,brockman2016openai} are a standard benchmark for state-of-the-art reinforcement learning algorithms and are difficult especially due to their high dimensional state and action spaces even though the physical systems they simulate are deterministic. We use code from \citep{fujimoto2018off} to recreate their ``imperfect demonstrations'' experiment set up, to see how the algorithms perform in these environments when faced with data generated by an expert RL policy that has been buried under lots of noise.

\begin{figure}[h]
    \center{
    \includegraphics[width=0.48\textwidth]
    {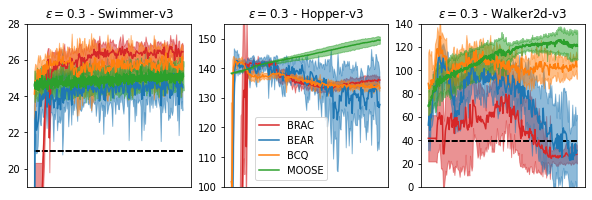}}
    \caption{\label{fig:mujoco} training curves in MuJoCo experiments---for each iteration we plot mean $\pm$ one standard deviation. MOOSE exhibits lower variance in the training process and leads to better policies than {  the other algorithms on Hopper and Walker}. In the Swimmer environment, {  it is outperformed by both BCQ and BRAC}. Dashed line represents original batch performance (about 89 for Hopper).}
\end{figure}

To this end, we train a DDPG agent through live interactions against the MuJoCo environments for 1,000,000 time steps and then have it act as an expert policy by collecting batches with 1,000,000 state-action-reward-next-state tuples through policy evaluation in the environment. We add Gaussian noise with a standard deviation of 0.3 to the action performed in 70\% of cases, and choose an action uniformly at random in the other 30\%. In order to investigate how MOOSE behaves when faced with various state and action dimensionalities, we perform experiments in the Swimmer-v3, Hopper-v3, and Walker2d-v3 environments, corresponding to state-action dimensionalities of 8 \& 2, 11 \& 3, and 17 \& 6.

\setlength{\tabcolsep}{1pt}
\begin{table}[h]
 
\centering
    \begin{tabular}{|ccc||ccrlcc|ccrlcc|ccrlcc|}
    \toprule
    \multicolumn{3}{|c||}{} & \multicolumn{6}{|c|}{Swimmer} & \multicolumn{6}{|c|}{Hopper} & \multicolumn{6}{|c|}{Walker2D} \\
    \midrule
    DDPG & & & & & -10.2 &$\pm$ 1.8 & & & & & 0.4 &$\pm$ 4.5 & & & & & -12.7 &$\pm$ 3.9& &\\
    BRAC & & & & & \textbf{25.9} &\textbf{$\pm$ 0.1} & & & & & 134.0 &$\pm$ 0.2 & & & & & 17.1 &$\pm$ 0.9& &\\
    BEAR & & & & & 22.9 &$\pm$ 0.1 & & & & & 117.4 &$\pm$ 1.3 & & & & & -7.0 &$\pm$ 3.8& &\\
    BCQ & & & & & 24.8 &$\pm$ 0.1 & & & & & 132.0 &$\pm$ 0.2 & & & & & 91.9 &$\pm$ 1.0& &\\
    MOOSE & & & & & 24.2 &$\pm$ 0.1 & & & & & \textbf{147.2} &\textbf{$\pm$ 0.2} & & & & & \textbf{113.7} &\textbf{$\pm$ 0.6}& &\\
    \bottomrule
    \end{tabular}
    
    \caption{Robust performance of the algorithms in the MuJoCo experiments. To assess robustness, 10\ts{th} percentile performance is shown together with its { standard error}. Final 10\% of performance values across { ten} seeds taken into account. MOOSE outperforms { its model-free counter parts on Hopper and Walker, while BRAC performs better on the Swimmer task.}}
    \label{table:mujoco}
\end{table}{}

\subsection{Industrial Benchmark}
The Industrial Benchmark \citep{hein2017benchmark} is a reinforcement learning benchmark environment motivated by industrial control problems, such as wind or gas turbines. Even though there is no single problem the benchmark tries to replicate, it exhibits problems commonly encountered in real-world industrial settings, such as high dimensional and continuous state
and action spaces, delayed rewards, complex noise patterns, and multiple counteracting objectives.\\
We generate batches of 100,000 data samples using the Industrial Benchmark with three different baseline policies:
\begin{itemize}
    \item The \textit{optimized} policy is an RL policy with very few parameters, taken from \citep{hein2018interpretable}. It was designed for interpretability, and found using a genetic algorithm (where policy assessment was performed via rollouts through models that received heavily explored data during training)
    \item The \textit{mediocre} policy is a   {simple, formula-based} policy which could stem from a human operator who has advised an automatic controller to keep the observable state variables at a fixed point. Its performance is worse than the optimized one, however still decent
    \item The \textit{bad} policy is also   {guiding the steerings to a fixed point}, however it was designed to be prohibitively bad in order to examine whether our algorithm can learn even from this kind of data.
\end{itemize}{}

\begin{figure}[h]
    \center{\includegraphics[width=0.48\textwidth]
    {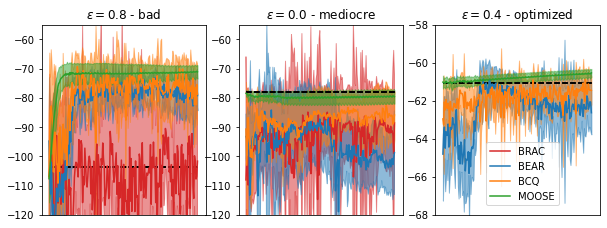}}
    \caption{\label{fig:performance} Mean performance $\pm$ one standard deviation over training time in the Industrial Benchmark experiments. MOOSE finds better performing policies while at the same time exhibiting lower variance over the course of training, which is important since no reliable policy selection techniques exist. Dashed line represents original batch performance. {  BRAC is omitted in the third plot due to too high variance - see Fig. \ref{fig:ib_full} for complete graphs.}}
\end{figure}

\setlength{\tabcolsep}{1.3pt}
\begin{table*}
  \begin{center}
   
    \begin{tabular}{|ccc|ccc|ccc||ccrclcc|ccrclcc|ccrclcc|ccrclcc|ccrclcc|ccrclcc|} 
    \toprule
       \multicolumn{9}{|c||}{$\varepsilon = $} & \multicolumn{7}{c|}{0.0} & \multicolumn{7}{c|}{0.2} & \multicolumn{7}{c|}{0.4} & \multicolumn{7}{c|}{0.6} & \multicolumn{7}{c|}{0.8} & \multicolumn{7}{c|}{1.0} \\
      \midrule
    & \parbox[t]{2mm}{\multirow{12}{*}{\rotatebox[origin=c]{90}{Behavior Baseline}}} & & & \parbox[t]{2mm}{\multirow{5}{*}{\rotatebox[origin=c]{90}{Bad}}} & & & DDPG & & & & -326.3 &$\pm$& 25.4 & & & & & -383.4 &$\pm$& 4.6 & & & & & -383.2 &$\pm$& 7.0 & & & & & -382.8 &$\pm$& 6.8 & & & & & -383.2 &$\pm$& 7.1 & & & & & & & & & \\
    & & & & & & & BRAC & & & & \textbf{-266.4} &\textbf{$\pm$}& \textbf{7.7} & & & & & -247.6 &$\pm$& 7.9 & & & & & -264.6 &$\pm$& 8.1 & & & & & -190.2 &$\pm$& 5.3 & & & & & -131.5 &$\pm$& 3.6 & & & & & & & & & \\
    & & & & & & & BEAR & & & & -322.4 &$\pm$& 3.1 & & & & & -180.3 &$\pm$& 3.5 & & & & & -129.4 &$\pm$& 2.7 & & & & & -98.3 &$\pm$& 1.29 & & & & & -88.73 &$\pm$& 0.92 & & & & & & & & & \\
    & & & & & & & BCQ & & & & -313.2 &$\pm$& 0.6 & & & & & -285.8 &$\pm$& 2.5 & & & & & -233.7 &$\pm$& 4.7 & & & & & -137.5 &$\pm$& 3.6 & & & & & -87.08 &$\pm$& 1.24 & & & & & & & & & \\
    & & & & & & & MOOSE & & & & -322.4 &$\pm$& 4.9 & & & & & \textbf{-125.9} &\textbf{$\pm$}& \textbf{2.2} & & & & & \textbf{-108.6} &\textbf{$\pm$}& \textbf{1.1} & & & & & \textbf{-90.26} &\textbf{$\pm$}& \textbf{0.26} & & & & & \textbf{-73.29} &\textbf{$\pm$}& \textbf{0.27} & & & & & & & & & \\

     \cline{4-44}
    & & & & \parbox[t]{2mm}{\multirow{5}{*}{\rotatebox[origin=c]{90}{Mediocre}}} & & & DDPG & & & & -959.9 &$\pm$& 33.8 & & & & & -960.3 &$\pm$& 35.8 & & & & & -960.3 &$\pm$& 37.4 & & & & & -960.6 &$\pm$& 24.4 & & & & & -960.1 &$\pm$& 48.5 & & & & & -178.0 &$\pm$& 3.3 & &\\
    & & & & & & & BRAC & & & & -114.3 &$\pm$& 1.9 & & & & & -103.0 &$\pm$& 1.5 & & & & & -112.1 &$\pm$& 1.8 & & & & & -98.97 &$\pm$& 2.63 & & & & & -101.1 &$\pm$& 2.7 & & & & & -106.8 &$\pm$& 2.1 & &\\
    & & & & & & & BEAR & & & & -111.7 &$\pm$& 1.2 & & & & & -103.0 &$\pm$& 4.6 & & & & & -123.7 &$\pm$& 3.2 & & & & & -104.6 &$\pm$& 3.2 & & & & & -99.75 &$\pm$& 1.6 & & & & & -66.31 &$\pm$& 0.22 & &\\
    & & & & & & & BCQ & & & & -103.3 &$\pm$& 1.1 & & & & & \textbf{-77.14} &\textbf{$\pm$}& \textbf{0.11} & & & & & \textbf{-72.71} &\textbf{$\pm$}& \textbf{0.35} & & & & & -75.86 &$\pm$& 0.74 & & & & & -106.7 &$\pm$& 2.4 & & & & & -69.16 &$\pm$& 0.39 & &\\
    & & & & & & & MOOSE & & & & \textbf{-82.95} &\textbf{$\pm$}& \textbf{0.28} & & & & & \textbf{-77.07} &\textbf{$\pm$}& \textbf{0.06} & & & & & -75.1 &$\pm$& 0.08 & & & & & \textbf{-70.53} &\textbf{$\pm$}& \textbf{0.18} & & & & & \textbf{-68.66} &\textbf{$\pm$}& \textbf{0.27} & & & & & \textbf{-64.27} &\textbf{$\pm$}& \textbf{0.03} & &\\
     
     \cline{4-44}
    & & & & \parbox[t]{2mm}{\multirow{5}{*}{\rotatebox[origin=c]{90}{Optimized}}} & & & DDPG & & & & -416.3 &$\pm$& 24.1 & & & & & -381.6 &$\pm$& 11.5 & & & & & -256.6 &$\pm$& 21.3 & & & & & -311.6 &$\pm$& 29.6 & & & & & -167.7 &$\pm$& 3.3 & & & & & & & & & \\
    & & & & & & & BRAC & & & & -113.4 &$\pm$& 2.6 & & & & & -77.32 &$\pm$& 0.75 & & & & & -158.2 &$\pm$& 6.0 & & & & & -89.21 &$\pm$& 1.74 & & & & & -115.6 &$\pm$& 2.6 & & & & & & & & & \\
    & & & & & & & BEAR & & & & -60.47 &$\pm$& 0.33 & & & & & -62.47 &$\pm$& 0.1 & & & & & -64.39 &$\pm$& 0.16 & & & & & -66.03 &$\pm$& 0.26 & & & & & -62.99 &$\pm$& 0.14 & & & & & & & & & \\
    & & & & & & & BCQ & & & & -60.12 &$\pm$& 0.04 & & & & & -60.86 &$\pm$& 0.07 & & & & & -62.71 &$\pm$& 0.11 & & & & & -63.56 &$\pm$& 0.21 & & & & & -72.38 &$\pm$& 0.49 & & & & & & & & & \\
    & & & & & & & MOOSE & & & & \textbf{-59.72} &\textbf{$\pm$}& \textbf{0.05} & & & & & \textbf{-60.35} &\textbf{$\pm$}& \textbf{0.02} & & & & & \textbf{-60.81} &\textbf{$\pm$}& \textbf{0.03} & & & & & \textbf{-62.07} &\textbf{$\pm$}& \textbf{0.01} & & & & & \textbf{-62.74} &\textbf{$\pm$}& \textbf{0.02} & & & & & & & & & \\
     
     \bottomrule
    \end{tabular}
    \caption{\label{table:performance}Performance of the algorithms on the Industrial Benchmark. In order to assess the algorithms’ robustness, the 10\ts{th} percentile performance is shown together with its { standard error}. Since no reliable offline policy selection techniques exist, the included policies stem from the final 10\% of training steps in experiments with {  ten} random seeds. MOOSE {  is outperformed only on 2 out of 16 datasets.}}
  \end{center}
\end{table*}

Each of the three baseline policies is then used for dataset generation in six exploration settings, in which we set the probability of performing a random action instead of the one recommended by the policy to be $\varepsilon = \{0.0, 0.2, 0.4, 0.6, 0.8,$ $ 1.0\}$. Altogether, these settings allow us to inspect the performance of our algorithm in the face of various issues encountered in real-world offline RL problems, i.e., optimized, human(-like), or undirected control strategies combined with various degrees of (possibly even no) exploration contained in the batch.

\section{Results \& Discussion}
We assess the performance of the algorithms by evaluating the trained policies in the original benchmark environment. As previously mentioned, robustness of the found solutions is critical, because we cannot hope to do policy selection much better than randomly (as opposed to supervised learning, where we can use the validation error to perform model selection). We thus examine the 10\ts{th} percentile performance instead of average performance. We find this to be a much more useful metric for practical applications, as we need to know what to expect from a worst-case perspective. In order to calculate it, we take all policies generated in the final 10\% of iterations across all random seeds into account. Results for the MuJoCo tasks are presented in Table \ref{table:mujoco}, and for the Industrial Benchmark experiments in Table \ref{table:performance}.\\

MOOSE outperforms { BRAC,} BEAR and BCQ in every experiment in terms of robust performance, except for one {  of the MuJoCo and two IB datasets}. Often, the margin by which MOOSE performs better is quite large. This highlights MOOSE's capability to produce better performing policies robustly over the course of training and across different initializations. Since reliable policy selection techniques (i.e., early stopping) are missing in offline RL \citep{hans2011agent}, this is an extremely important aspect. Compared to {  the model-free algorithms'} high variance (see Fig. \ref{fig:performance} and \ref{fig:mujoco}), MOOSE's training curves are much smoother, producing well performing policies more robustly. It thus better facilitates application in true offline RL settings. As expected, DDPG performs worst in all experiments, except for one, where it (randomly) manages to improve upon the prohibitively bad baseline in the absence of any exploratory moves ($\varepsilon = 0.0$). Note that we exclude DDPG's results in Figures \ref{fig:mujoco}, \ref{fig:performance}, and \ref{fig:ib_full} and only show them in Tables \ref{table:mujoco} and \ref{table:performance} for better comparability of the other algorithms. \\
Given the experimental results, we find that MOOSE can learn purely from batch data in various continuous control environments, even when faced with high dimensional state and action spaces, delayed rewards, or complex noise. Since throughout all policy and exploration settings in our experiments, MOOSE proves to be an effective learning algorithm, we find it can be trusted when faced with   {various kinds of} policies, both narrow and wide data distributions, as well as undirected or suboptimal data generating agents.\\
{ Noteworthy is the performance of BRAC: While it does better than most other algorithms on Swimmer and Hopper, it severely underperforms on the industrial benchmark. We hypothesize, that both the representation of the behavior policy (a transformed Gaussian - which does not fit to the $\varepsilon$-greedy approach of data generation), as well as the sample based estimation of the KL divergence contribute to its large variance and thus its lower robustness.}\\
A further advantage of MOOSE over BCQ is that it finds a closed form policy: The algorithm thus scales better with its hyperparameters than BCQ, since it does not rely on sampling from a generative model. If we would like to decrease the penalty on the policy because we observe that it is keeping too strictly to the original policy, we can simply decrease the corresponding weight in Equation \ref{eq:final_loss}.\\
Finally, MOOSE produces deterministic policies, which can be a requirement in real-world applications due to safety concerns. BEAR's and BCQ's approaches are dependent on stochastic policies, so MOOSE constitutes a step towards better applicability also in this dimension.

\section{Conclusion}
In this paper, we introduced MOOSE, a novel model-based reinforcement learning algorithm, designed specifically for the offline RL setting, that constrains its policy directly to be close to the previously collected batch data, without detours through model uncertainty. We also do not strictly constrain towards replication of the behavior policy, but rather for the trained policy to have support in the original dataset for the states visited and actions chosen in (imagined) trajectories. We compared our algorithm with state-of-the-art model-free algorithms for the offline RL setting: { BRAC,} BEAR and BCQ. We find that model-based RL can play to its strengths in the offline setting, since it makes more effective use of the limited amount of data available. Furthermore, MOOSE produces much more stable results than the compared model-free methods, which is a key requirement in offline reinforcement learning due to the lack of a reliable offline policy selection technique. We find that MOOSE outperforms its model-free counterparts in almost all considered cases, often even by far.

\section*{Acknowledgements}
The project this paper is based on was supported with funds from the German Federal Ministry of Education and Research under project number 01\,IS\,18049\,A.
\newpage

\bibliography{main}
\bibliographystyle{apalike}

\clearpage

\appendix

\section{Experimental Details}
\subsection{Hyperparameters}
In this paper, we did not tune or change hyperparameters throughout experiments. All neural network transition models or policies have two hidden layers of size 400 and 300 with ReLU activation functions. The transition models have normalized weights and no nonlinearity in their final layer, while policies end with a ${\rm tanh()}$, since we assume actions to lie in $(-1, 1)$. Variational Autoencoders also use ReLUs, have one layer of size 750, two parallel layers (one for the mean, one for the variance) of size $2*{\rm actiondim}$ for the parameterization of the latent variables, and then another two layers of size 750 after the bottleneck. All transition and autoencoder networks are trained with a batch size of 500, while policies are trained with 100 start states per gradient step. For autoencoder and transition (or reward) models we use the Adam optimizer with a learning rate of $10^{-4}$ and standard hyper parameters. The policy networks for the MuJoCo experiments had a lower learning rate of $10^{-5}$. In the Industrial Benchmark experiments, we observed that the momentum style components of Adam hurt the optimization and we resorted to vanilla SGD with a learning rate of $10^{-4}$. To avoid gross extrapolation mistakes, we clip model predictions to stay inside the range of values that have been observed in the batch.\\
During evaluation on the real benchmark environments, we always averaged performance over 10 trajectories of length 100. Consequently, we also used a horizon of 100 during policy training. In the offline RL setting, there is no real way to know when to stop learning, we thus pick a reasonable number of iterations to train and stick to it. For the MuJoCo tasks, we train { BRAC,} BEAR and BCQ for 100,000 steps, while we train the models and autoencoders in MOOSE separately from the policy for 50 epochs, and the policy for 5000 steps. We deviate from this behavior only in the Hopper environment, as the transition models start to predict premature falling over of the Hopper after about 1000 steps. Model-based algorithms cannot take into account the length of the trajectories in the same way that model-free approaches can, since they are unable to differentiate this attribute (at least not without some additional engineering overhead). In the IB experiments we decrease training steps of the policies in all three algorithms, since the dataset size also decreased to 100,000 samples. { BRAC,} BEAR and BCQ are trained for 10,000 and MOOSE for 1,000 steps.\\
The $\lambda$ parameter that controls the tradeoff between optimizing the policy for best possible return and for closeness to the original data distribution was always left at a conservative $0.01$ ($1-\lambda = 0.99 $), meaning that we mostly focused on staying close to the data (both reward and penalty are computed in a space normalized to have zero mean and unit standard deviation). The choice of hyper parameters in offline RL is an especially hard problem, since we cannot know à priori how close is close enough to the data. We thus find being conservative the only viable option in practice. The $\eta$ parameter controlling the tradeoff between optimizing for average performance versus worst case performance was always kept at $0.5$. Throughout experiments, we use $K=4$ transition (and reward) models to estimate the rewards across trajectories. In accordance with previous literature we use a discount factor $\gamma=0.99$ for the MuJoCo and $\gamma=0.97$ for the Industrial Benchmark. Since the setpoint parameter in the Industrial Benchmark simulates ambient conditions out of control of the learner and we do not aim to perform transfer learning, we do not alter it and instead keep it fixed at $p=70$.
\subsection{Prior Knowledge}
As prior results have shown that learning deltas of the transitions instead of directly predicting next states can be beneficial for the MuJoCo tasks, we use delta models in those experiments. In the Industrial Benchmark experiments, we found that delta models performed worse on the held out evaluation trajectories than models that directly predicted successor states. We hypothesize that the delta models in this case are harder to learn due to the rather noisy transitions. Hence, we used models that directly predicted successor states throughout Industrial Benchmark experiments.\\
The three steerings velocity, gain, and shift in the Industrial Benchmark are always updated by the chosen control action times a steering specific constant. We assume that this is low level domain expert knowledge that could be handily available in a real world setting and directly integrate this in our model building process. As a consequence, our models need to only predict the other parts of the state space. Since the model-free methods do not predict transitions, they cannot benefit from this prior knowledge.

\section{Uncertainty Calculation for 10\ts{th} percentiles}
To calculate uncertainties for the 10\ts{th} percentile performances, the naïve way would be to repeat the entire series of experiments (we performed each experiment { ten} times) another $J$ times and calculate the uncertainty based on that. As that would be wasteful, and since we work with limited computational resources, we work with the data that we already have:\\

\begin{itemize}[noitemsep,topsep=0pt]
    \item we take all policy performance values that we already took into account for the percentile calculation (final 10\% of iterations) and make the assumption that they follow a Gaussian distribution (visualizations show this is justified, even though strictly they are not independent).
    \item we calculate the standard error of the mean of the policy performance values and multiply it by $1.7$ as we find through Monte Carlo experiments that the 10\ts{th} percentile value is roughly $1.7$ times as uncertain as the mean, when the underlying data is normally distributed.
\end{itemize}

\section{Full Figures for Experiments on the Industrial Benchmark}
\begin{figure*}[!htbp]
    \center{\includegraphics[width=0.95\textwidth]{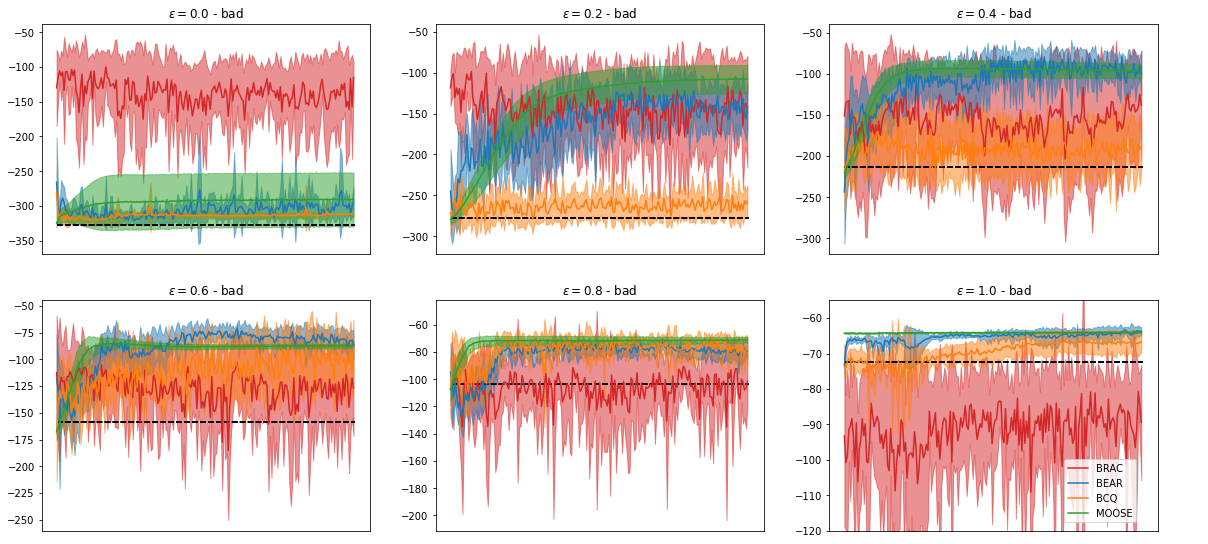}}
    \center{\includegraphics[width=0.95\textwidth]{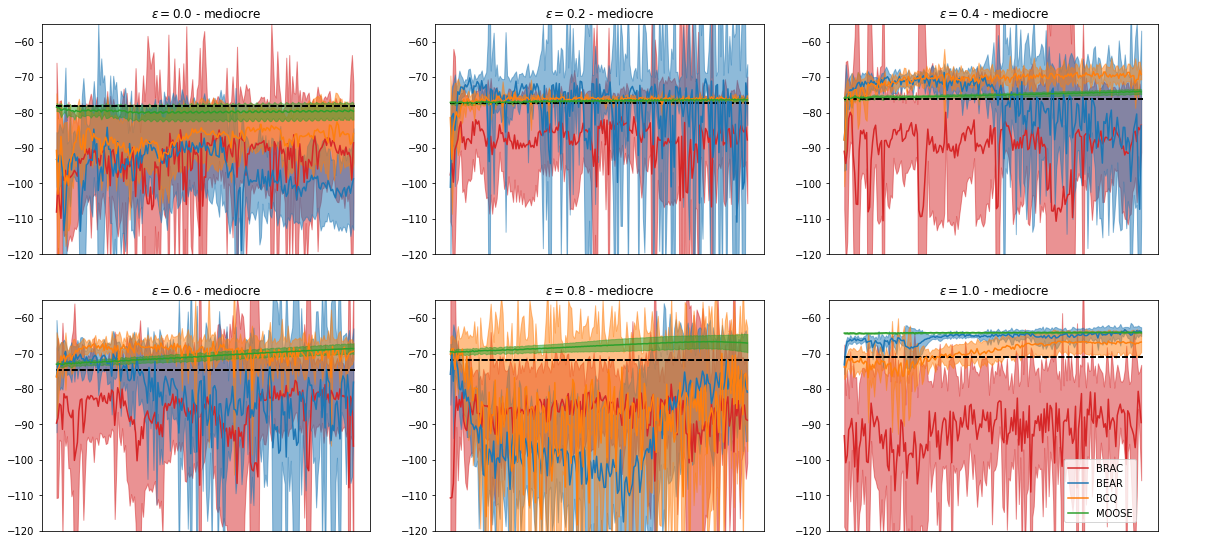}}
    \center{\includegraphics[width=0.95\textwidth]{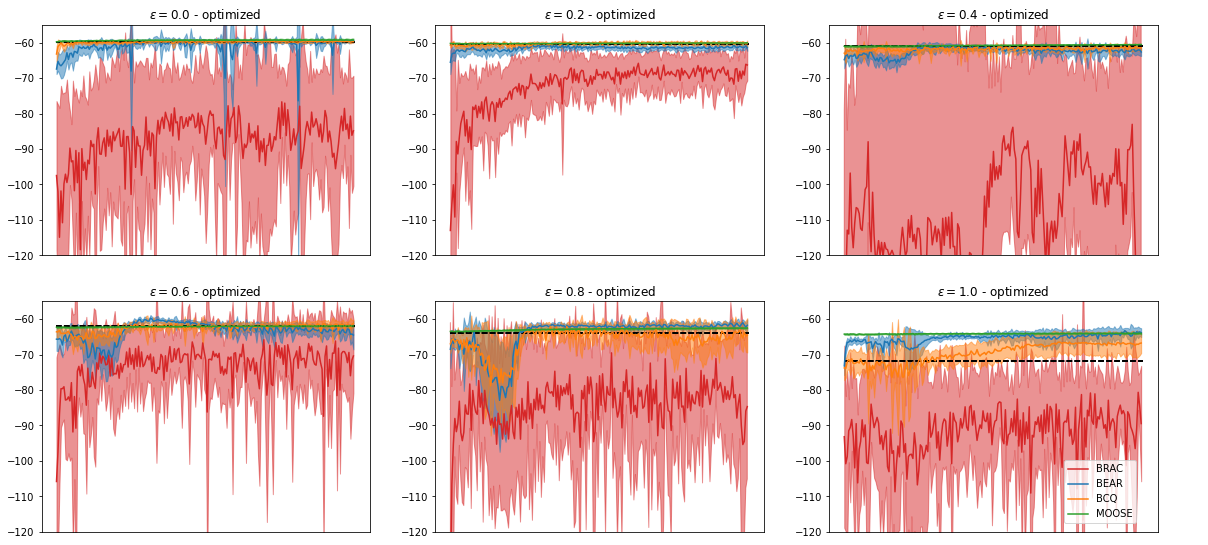}}
    \center{\caption{\label{fig:ib_full}Mean performance $\pm$ standard deviation in IB experiments. Dashed line represents original batch performance.}}
\end{figure*}

The complete graphs for the industrial benchmark experiments are shown in Figure \ref{fig:ib_full}.

\onecolumn
\section{Detailed Derivation of Equation \ref{eq_minimize}}
\label{app_eq_minimize}
We provide a detailed derivation of Eq. \ref{eq_minimize}, which states that minimizing the expected model bias of a newly derived policy with parameters $\theta$ is equivalent to maximizing the probabilities of the states it visits and the actions it chooses under the original dataset.
\begin{eqnarray}
    \mathbf{\theta^*} = &\argmin\limits_\theta& \mathbb{E}_{s \sim \mu_{\beta}^e,\; a \sim \beta(s),\; \hat{s} \sim \mu_{\pi_{\theta}}^f,\; \hat{a} \sim \pi_{\theta}(\hat{s})} \left[B\right]\\
    = &\argmin\limits_\theta& \mathbb{E}_{s \sim \mu_{\beta}^e,\; a \sim \beta(s),\; \hat{s} \sim \mu_{\pi_{\theta}}^f,\; \hat{a} \sim \pi_{\theta}(\hat{s})} \left[\sum_t b_t\right]\\
    = &\argmin\limits_\theta& \mathbb{E}_{s \sim \mu_{\beta}^e,\; a \sim \beta(s),\; \hat{s} \sim \mu_{\pi_{\theta}}^f,\; \hat{a} \sim \pi_{\theta}(\hat{s})} \left[\sum_t  \mathbb{E}_{s_{t+1} \sim e(s_t,a_t),\; \hat{s}_{t+1}=f(\hat{s}_t,\hat{a}_t)} \left[||s_{t+1} - \hat{s}_{t+1}||_2\right]   \right]\\
    = &\argmin\limits_\theta& \mathbb{E}_{s \sim \mu_{\beta}^e,\; a \sim \beta(s),\; \hat{s} \sim \mu_{\pi_{\theta}}^f,\; \hat{a} \sim \pi_{\theta}(\hat{s}),\; t\in T} \left[\mathbb{E}_{s_{t+1} \sim e(s_t,a_t),\; \hat{s}_{t+1}=f(\hat{s}_t,\hat{a}_t)} \left[(s_{t+1} - \hat{s}_{t+1})^2\right]\right] \label{final_eq}
\end{eqnarray}
We make the assumption that the distribution of model errors $e(s, a) - f(\hat{s}, \hat{a})$ has a variance that is monotonically decreasing with the likelihood of having seen the imagined data in reality, i.e., that the data sample $(\hat{s}, \hat{a})$ was generated under the original environment dynamics $e$ and the behavior policy $\beta$:
\begin{equation}
    (s' - \hat{s'}) \sim \mathcal{N}(0, -\log p_{e,\beta}(\hat{s}, \hat{a}))
\end{equation}
Since the expectation of a squared Gaussian variable with zero mean is its variance\ts{*}, plugging the assumption into Eq. \ref{final_eq} yields
\begin{eqnarray}
    \mathbf{\theta^*}= &\argmin\limits_\theta& \mathbb{E}_{\hat{s} \sim \mu_{\pi_{\theta}}^f,\; \hat{a} \sim \pi_{\theta}(\hat{s}),\; t\in T} \left[-\log p_{e,\beta}(\hat{s}_t, \hat{a}_t)\right]
\end{eqnarray}{}
Minimizing the negative log likelihood of having observed the imagined data in reality then corresponds to maximizing the actual likelihood as given by the second part of the original Eq. \ref{eq_minimize} (expectation subscript omitted for brevity):\\
\begin{eqnarray}
    \mathbf{\theta^*} = &\argmin\limits_\theta& \mathbb{E} \left[-\log p_{e,\beta}(\hat{s}, \hat{a})\right] \\
    = &\argmax\limits_\theta& \mathbb{E} \left[\log p_{e,\beta}(\hat{s}, \hat{a})\right] \\
    = &\argmax\limits_\theta& \mathbb{E} \left[p_{e,\beta}(\hat{s}, \hat{a})\right] \\
    = &\argmax\limits_\theta& \mathbb{E} \left[\mu_{\beta}^e(\hat{s}) \beta(\hat{a}|\hat{s})\right]
\end{eqnarray}{}

* Expectation of a squared Gaussian variable $X \sim \mathcal{N}(0, s^2)$ can easily be derived from the formula for its variance:

\begin{eqnarray}
    &\mathbb{V}[X] =& \mathbb{E}[X^2] - (\mathbb{E}[X])^2 \\
    \Leftrightarrow &\mathbb{V}[X] =& \mathbb{E}[X^2] - 0^2 \\
    \Leftrightarrow &\mathbb{E}[X^2] =& s^2
\end{eqnarray}

\end{document}